\documentclass[article,12pt]{elsarticle}

\usepackage{amssymb}
\usepackage{pdflscape}
\usepackage{multirow}
\usepackage{array}
\usepackage{url}
\usepackage{amsmath}
\usepackage{amssymb}
\usepackage{textcomp}
\newcommand{\textapprox}{\raisebox{0.5ex}{\texttildelow}}

\journal{Image and Vision Computing}

\newcommand{\eg}{{\it e.g.},~}
\newcommand{\ie}{{\it i.e.},~}

\begin{document}

\begin{frontmatter}

\title{Implications of Ocular Pathologies \\for Iris Recognition Reliability}

\author[addressNASK,addressPW]{Mateusz Trokielewicz}

\author[addressNASK,addressPW]{Adam Czajka}

\author[addressWUM]{Piotr Maciejewicz}

\address[addressNASK]{Biometrics Laboratory, Research and Academic Computer Network (NASK)\\Kolska 12, 01-045 Warsaw, Poland}
\address[addressPW]{Institute of Control and Computation Engineering, Warsaw University of Technology\\Nowowiejska 15/19, 00-665 Warsaw, Poland}
\address[addressWUM]{Department of Ophthalmology, Medical University of Warsaw\\Lindleya 4, 02-005 Warsaw, Poland}

\begin{abstract} 
This paper presents an analysis of how iris recognition is influenced by eye disease and an appropriate dataset comprising 2996 images of irises taken from 230 distinct eyes (including 184 affected by more than 20 different eye conditions). The images were collected in near infrared and visible light during routine ophthalmological examination. The experimental study carried out utilizing four independent iris recognition algorithms (MIRLIN, VeriEye, OSIRIS and IriCore) renders four valuable results. First, the enrollment process is highly sensitive to those eye conditions that obstruct the iris or cause geometrical distortions. Second, even those conditions that do not produce visible changes to the structure of the iris may increase the dissimilarity between samples of the same eyes. Third, eye conditions affecting the geometry or the tissue structure of the iris or otherwise producing obstructions significantly decrease same-eye similarity and have a lower, yet still statistically significant, influence on impostor comparison scores. Fourth, for unhealthy eyes, the most prominent effect of disease on iris recognition is to cause segmentation errors. To our knowledge this paper describes the largest database of iris images for disease-affected eyes made publicly available to researchers and offers the most comprehensive study of what we can expect when iris recognition is employed for diseased eyes.
\let\thefootnote\relax\footnote{Submitted for the Image and Vision Computing Journal on 6 October 2015, revised on 2 july 2016, accepted for publication on 14 August 2016.}
\end{abstract}


\end{frontmatter}

\section{Introduction}
\label{introduction}

\subsection{Iris recognition in the presence of ocular pathologies} 

The iris is a colored, annular structure surrounding the pupil of the eye. It displays intrinsically complex patterns that are unique due to random epigenetic factors. The distinctive patterns, developed in the fetal stage, provide features that can be used with high confidence for biometric identification. Uniqueness, combined with the feasibility of fast template creation and matching, allows for the building of large-scale applications of iris recognition. Examples are India’s AADHAAR Unique Identification Authority of India \cite{AADHAAR} with more than 1.2 billion people enrolled, and CANPASS system \cite{CBSA} maintained by the Canadian Border Services Agency (CBSA) which provides efficient entry into Canada for frequent travelers. 

Iris recognition is also being considered by the International Civil Aviation Organization (ICAO) as a candidate for inclusion in the next generation of biometric passports along with fingerprint and face biometrics \cite{ICAO}. The temporal stability of iris patterns is strongly supported by Safir and Flom in a 1987 patent describing theoretical principles of iris recognition. The authors assert that the \emph{'significant features of the iris remain extremely stable and do not change over a period of many years'} \cite{SafirFlom}. Further support comes from John Daugman whose 1994 patent states: \emph{'the iris of the eye is used as an optical fingerprint, having a highly detailed pattern that is unique for each individual and stable over many years'} and \emph{'the iris of every human eye has a unique texture of high complexity, which proves to be essentially immutable over a person’s life'} \cite{DaugmanPatent}. The NIST IREX VI report presents an experimental study showing lack of aging in the iris \cite{IREX6}. (However, one should be aware of other studies presenting evidence of aging in the iris, published by University of Notre Dame, USA \cite{Fenker2011,Fenker2012,Baker2013}, and Warsaw University of Technology, Poland \cite{Czajka2014,TrokielewiczAgingIWBF2015}.) 

Those statements and the excellent performance of iris recognition are related to the immutability of the iris pattern in a healthy eye. When an injury, disease or other ocular pathology is present, it is not a given that such a condition will not affect the visibility or appearance of the features of the iris. Such occurrences may degrade the performance of the iris recognition system either by altering the iris or obstructing its view. In certain circumstances, it may even render the eye unsuitable for use in authentication at all. This can be the case in occurrences of aniridia, a medical disorder, often bilateral, in which only a small, ring-shaped portion of tissue is present where the iris would normally be located. This often leaves a large and possibly irregularly shaped pupil \cite{Aniridia}.

Eye trauma and injuries are also contributing factors to degradation of iris recognition accuracy. Although Daugman states that \emph{'as an internal organ of the eye the iris is well protected from the external environment'} \cite{DaugmanPatent}, in some cases this kind of protection (\ie by the cornea and the aqueous humor in the anterior chamber of the eye) may not be sufficient. Safir and Flom, being ophthalmologists and thus aware of this issue, acknowledged it in their patent claim: \emph{'A sudden or rapid change in such a feature [of iris pattern] may result in a failure to identify an individual, but this may alert the individual to the possibility of pathology of the eye'} \cite{SafirFlom}.

\subsection{Scope of this study} 

Various medical conditions affecting the structures of the eye, the iris in particular, may cause a deterioration of the reliability of iris recognition. Investigation into previous research (cf. Sec. \ref{related}) shows that we are still far from fully understanding how various eye conditions impact iris recognition. This can be attributed to the lack of large, heterogeneous, and publicly available databases appropriate for this subject. This paper aims to answer \textbf{four questions} related to ocular disorders and their impact on iris recognition while providing an appropriate database of images of unhealthy eyes to researchers. \textbf{The questions are:}

\begin{enumerate}
\item Do ocular pathologies impact the enrollment process? If so, which structural impairments translate into an increase in the Failure to Enroll rate (FTE), \ie the proportion of samples that could not be enrolled to the overall number of samples?
\item Does iris recognition perform worse in unhealthy eyes without visible impairments in comparison with healthy irises when photographed in near-infrared (NIR) light? In other words, can we assume that there are some properties of the iris image, not revealed in NIR light, that prevent iris recognition from achieving optimal performance?
\item What kind of visible impairments in unhealthy irises have the greatest impact on iris recognition?
\item What are the main reasons for bad performance when iris recognition is applied to unhealthy eyes?
\end{enumerate}

To answer these questions, a dataset of iris images representing more than twenty different eye diseases was built with the use of a professional iris recognition camera operating in NIR light, along with an ophthalmological commentary (cf. Sec. \ref{database}). Most of the NIR samples are accompanied by color images to make possible independent ophthalmological interpretations. Experimental study done for four different and independent iris recognition algorithms is presented (cf. Sec. \ref{section:experimental} and \ref{experiments}).

To our knowledge, this paper describes the largest published dataset of NIR and color images for unhealthy eyes with a professional, ophthalmological commentary and offers the most extensive study to date of ways in which different types of diseases of the eye impact iris recognition.

\section{Related work}
\label{related}

Previous studies related to the topic of the influence of ocular disease on iris recognition show that the need for large, heterogeneous, publicly available databases has yet to be fully appreciated.  With few exceptions, most heretofore available studies utilize small datasets and focus on only one disease.

Probably the first experiment devoted to this field of research was conducted by Roizenblatt \emph{et al.} \cite{Roizenblatt} and involved 55 patients suffering from cataract. Each person was enrolled in the LG IrisAccess 2000 biometric system before cataract removal surgery, 30 days after cataract removal, and 7 days after stopping the administration of pupil-dilating drugs (when pupils had reverted to normal size and reaction to light). After the 30-day period, differences in the size of pupils  were no larger than 1.5mm when compared with images obtained prior to the treatment. Each eye that underwent a surgery was also given a score between 0 and 4 based on visual inspection performed by an ophthalmologist. One point was given for each of the following ocular pathologies: depigmentation, pupil ovalization, focal atrophy with and without trans-illumination. A correlation was revealed between the visual inspection score and change in Hamming distances (HD) between templates created using pre- and post-surgery samples. 6 out of 55 eyes were no longer recognized, thus yielding FNMR of about 11\%. For remaining irises there were significant shifts towards worse HD (11.3\% increase in average HD when scores between gallery samples and post-surgery samples are compared against scores between gallery samples and pre-surgery samples) and worse visual scores (11.13\% increase in average visual score for images collected post-surgery, as compared to average visual score for those obtained pre-surgery), however, those eyes were classified correctly. Authors suggest that the energy released inside the eyeball during the cataract surgery may be a cause of atrophic changes to the iris tissue. Re-enrollment is suggested as a countermeasure in cases with significant, visual alteration to the iris visible during a slit-lamp examination.

A similar scenario regarding the impact of cataract surgery was explored by Seyeddain \emph{et al.} \cite{Seyeddain2014} who performed an experiment to establish the effects of phacoemulsification and pharmacologically induced mydriasis on the iris. (Phacoemulsification refers to the extraction of the lens through aspiration. The procedure involves the insertion of a small probe through an incision in the side of the cornea. The probe emits ultrasound waves that break up the opacificied lens which is later removed using suction \cite{TrokielewiczWilga2014}.) The experiment aimed to determine whether the irises, following phacoemulsification or drug induced mydriasis (preventing the dilated pupil from reacting to light stimulation) perform more poorly when compared to the same irises before the procedure or before the drug-induced pupil dilation. They revealed that 5.2\% of the eyes that were subject to cataract surgery could no longer be recognized after the procedure. In the pupil dilation group, this portion reached as high as 11.9\%. In both cases the authors suggest re-enrollment for patients whose eyes were not successfully identified after the surgery or instillation of mydriatics. No false acceptances were observed in either case.

Trokielewicz \emph{et al.} \cite{TrokielewiczWilga2014} aimed at quantifying the impact of cataracts on iris recognition performance. An experiment involving three different iris recognition methods revealed differences in system performance when comparison scores calculated before surgery from cataract-affected eyes are used instead of those obtained from healthy eyes. For all three methods there was a perceivable degradation in average genuine comparison scores with differences reaching from 12\% of genuine score increase for an academic matcher, up to 175\% of genuine score increase for one of the commercial matchers. For two out of three matchers, these changes also affected the final false non-match rate.

Dhir \emph{et al.} \cite{Dhir} studied the influence of the effects of mydriatics accompanying cataract surgery. A group of 15 patients had their eyes enrolled before surgery. A verification was performed at 5, 10, and 15 minute intervals after application of the drug and again two weeks following the procedure itself. None of the eyes was falsely rejected after this two-week period.  One must, however, keep in mind that the authors excluded from the dataset eyes with pre-existent corneal and iris pathologies, or those with iris tissue damaged during the surgery. The study suggests that recognition deterioration may originate from a slight shift of the iris towards the center of the eyeball resulting from implantating an artificial lens that is thinner than the natural lens. Specular reflections from the implant may also contribute to erroneous segmentation. However, increase in pupil diameter induced by mydriatics led to FNMR of 13.3\%, as 6 out of 45 verification attempts failed. In addition, Hamming distances increased with the elapse of time after the instillation of the drug. The authors warn that this phenomenon may be exploited by criminals in order to enroll in a biometric system under multiple identities to deceive law enforcement organizations.

Yuan \emph{et al.} \cite{Yuan} examine another type of medical procedure -- laser-assisted refractive correction surgery -- and its possible impact on iris recognition. These procedures take advantage of laser radiation to ablate the corneal tissue and compensate for refractive pathologies such as myopia, hypermetropia and astigmatism. Researchers carried out an experiment to find out whether such manipulation may result in increased FNMR of an iris biometric system. Using Masek’s algorithm for encoding, 13 eyes (out of 14) were correctly recognized after a procedure had been performed. However, the one eye that was falsely non-matched had a significant deviation in circularity of the pupil and increased pupil diameter. Therefore, the authors argue that refractive correction procedures have little effect on iris recognition. Nonetheless, further experiments incorporating larger datasets are called for.

In a study by Aslam \emph{et al.} \cite{Aslam}, 54 patients suffering from several different eye conditions were enrolled in a biometric system using the IrisGuard H100 camera during their first visit. Their eyes were again photographed after treatment. Researchers calculated Hamming distances between iris images (encoded using Daugman’s algorithm) obtained before and after the treatment to determine whether treatment had any impact on recognition accuracy. Tested methodology turned out to be resilient for most illnesses, \ie glaucoma treated using laser iridotomy, infective and non-infective corneal pathologies, episcleritis, scleritis and conjunctivitis. However, 5 out of 24 irises affected by the anterior uveitis (a condition in which the middle layer of the eye, the uvea, which includes the iris and the cilliary body, becomes inflamed \cite{Uveitis}) were falsely non-matched after treatment, producing an FNMR rate of about 21\% (with acceptance threshold set to $HD=0.33$). It is worth noting that each of the eyes that yielded a false non-match had earlier been administered mydriatics; therefore, the pupil was significantly dilated. In addition, two eyes suffered from high corneal and anterior chamber activity, while the remaining three had posterior synechiae that caused deviation from the pupil circularity.  The hypothesis stating that the mean Hamming distance in the anterior uveitis subset is equal to that of the control group (consisting of healthy eyes) has been rejected with $p < 10^{-4}$, while there were no statistically significant differences between scores obtained from other disease subsets when compared to the control group (thus the null hypotheses could not be rejected). As for the pathologies related to the corneal opacities, Aslam tries to explain lack of recognition performance deterioration by the fact that the NIR illumination used in iris biometrics is more easily transmitted through such obstructions and therefore allows correct imaging of underlying iris details. Laser iridotomy also showed little influence, as the puncture in the iris tissue made by laser radiation appears to be too small to significantly alter the iris pattern. However, certain combinations of synechiae and pupil dilation can affect the look of the iris texture sufficient to produce recognition errors. A deviation in the pupil’s circularity caused by the synechiae may also contribute to segmentation errors.

Borgen \emph{et al.} \cite{Borgen} conduct a study focusing on iris and retinal biometrics in which they take advantage of 17 images selected from the UBIRIS database. Those images were then digitally modified to resemble changes to the eye structures caused by various ocular illnesses: keratitis and corneal infiltrates, blurring and dulling of the cornea, corneal scarring and surgery, angiogenesis, tumors and melanoma. High FNMR values (32.8\% -- 86.8\%) are reported for all modifications except for the pathological vascularization (6.6\%), changes in  iris color (0.5\%) and iridectomy-derived damage in the iris, for which FNMR=0. Faulty segmentation is suspected to be the main cause, especially in cases involving clouding of the cornea. The authors, however, do not acknowledge the fact that NIR illumination enables correct imaging even in eyes with corneal pathologies such as clouding or other illness-related occlusions.

In a study by McConnon \emph{et al.} \cite{McConnon2012} three groups of medical disorders (conditions causing pupil/iris deformation, pupil/iris occlusion and eyes having no iris or a very small iris) were distinguished to estimate the impact they may have on the reliability of iris segmentation. Due to lack of publicly available datasets, the database used in this work consisted of images drawn from the Atlas of Ophthalmology, making them imperfectly suited for iris recognition (\ie for having been captured in visible light). Those images have been resampled to $320\times240$ resolution and manually segmented to obtain the ground truth iris localization. Automatic segmentation, performed using Masek’s algorithm, deviated by two or more pixels in 46\% and 55\% of images for the limbic and pupillary boundaries, respectively.

Our most recent work expands our earlier experiments regarding cataract-related effects and attempts to assess which types of eye damage caused by disease have the greatest impact on the accuracy of a given biometric system \cite{TrokielewiczCYBCONF2015}. In this preliminary study we have shown that changes to the fabric of the iris and geometrical distortions in the pupillary area have the highest potential of degrading genuine comparison scores for three different iris recognition algorithms employed in that research. This study was later expanded even further \cite{TrokielewiczBTAS2015} with data collected from more ophthalmology patients over a longer period of time, as well as with experiments that revealed an increased chance of failure-to-enroll errors when iris biometrics systems are presented with images obtained from patients with diseased eyes. We also pointed to segmentation stage errors as the most probable source of deteriorated matching accuracy. The datasets of eye images affected by ocular disorders, used in the preliminary and the expanded research, are publicly available to all interested researchers.

This paper offers a significant extension of the work presented in \cite{TrokielewiczBTAS2015}, including an analysis for the fourth, additional iris recognition algorithm. To our best knowledge this study presents the most comprehensive and up-to-date insight into the subject of the influence of ocular pathologies on the reliability of iris biometrics.

\section{Database of iris images}
\label{database}

\subsection{Medical background}

The successful performance of an iris biometric system depends on the capacity to correctly image details of the texture of the iris. Imaging of irises afflicted by diseases can be problematic for several reasons. Heavily distorted pupils, deviating severely from their usual circular shapes, can affect image segmentation algorithms approximating pupillary and limbic boundaries with circles. Eyes with changes to the cornea may perform worse when segmentation methods utilizing image local gradients are used. Severe damage to the iris tissue itself may change the pattern significantly, sufficient to make correct identification impossible. The following subsection presents a brief characterization of medical conditions represented in the database created for this study. The potential impact on the performance of a biometric system, depending on the category of these disorders, is discussed as well.

\paragraph{\textbf{The cornea}} As the outermost part of the eye, the cornea, despite being fairly durable, can still suffer from numerous factors. Chemical injury can deal extensive damage to the ocular surface epithelium, the cornea, and the anterior segment of the eye. It can lead to \textbf{opacification and neovascularization of the cornea}, formation of a \textbf{symblepharon and cicatricial ectropion or entropion}. If significant \textbf{corneal scarring} is present, a corneal transplant may be required to restore vision. A benign growth of the conjunctiva -- \textbf{pterygium} -- commonly forms from the nasal side of the sclera to the center of the cornea. This fibrovascular proliferation often occludes a part of the iris. \textbf{Bacterial keratitis} is an erosion or an open sore in the outer layer of the cornea with stromal infiltration, edema and hypopyon. Common pathogens that may lead to \textbf{corneal ulcers} include Streptococcus pyogenes, Acanthamoeba, Herpes simplex, or fungal infections mainly caused by use of non-sterilized contact lenses. \textbf{Acute glaucoma}, with increased pressure inside the eye, can occur suddenly when the iris is pushed or pulled forward. High intra-ocular pressure produces symptoms such as \textbf{corneal edema, shallowness of the anterior chamber and dilatation of the pupil} which may become oval in shape. \textbf{Corneal laceration} usually requires placement of corneal sutures. These disorders usually impact the look and clarity of the corneal area, partially or totally covering the iris.

\paragraph{\textbf{The anterior chamber}} \textbf{Hyphema} is a condition characterized by the presence of blood in the anterior chamber of the eye that can partially or completely obstruct the view of the iris. Hyphemas are frequently caused by injuries but may also occur spontaneously. A long-standing hyphema may result in hemosiderosis and heterochromia in a form of partial changes to iris coloration. \textbf{Hypopyon} is a leukocytic exudate present in the anterior chamber of the eye, usually accompanied by redness of the conjunctiva. It is a sign of an iridial inflammation. Both conditions may significantly obstruct the view of the iris by obscuring the iris texture, thus causing problems with segmentation.

\begin{figure}[!htb]
\centering
\includegraphics[width=0.492\textwidth]{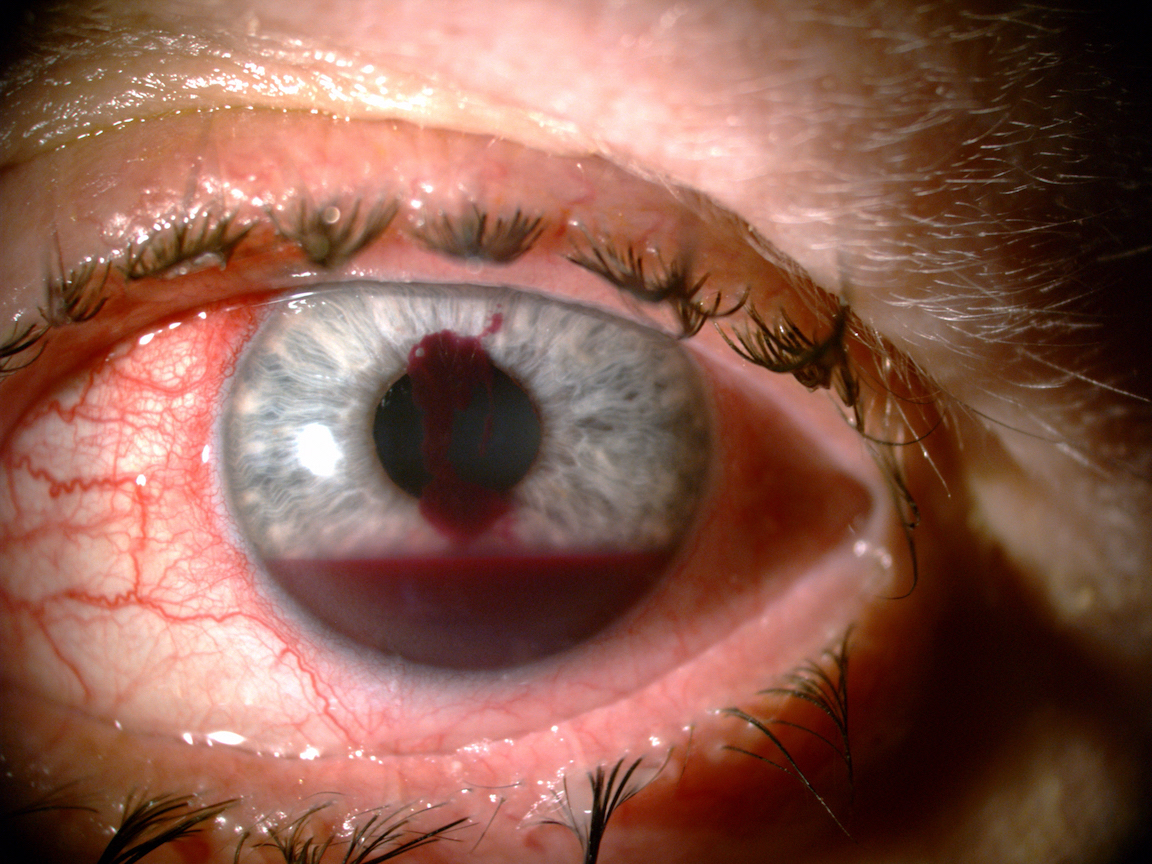}\hfill
\includegraphics[width=0.492\textwidth]{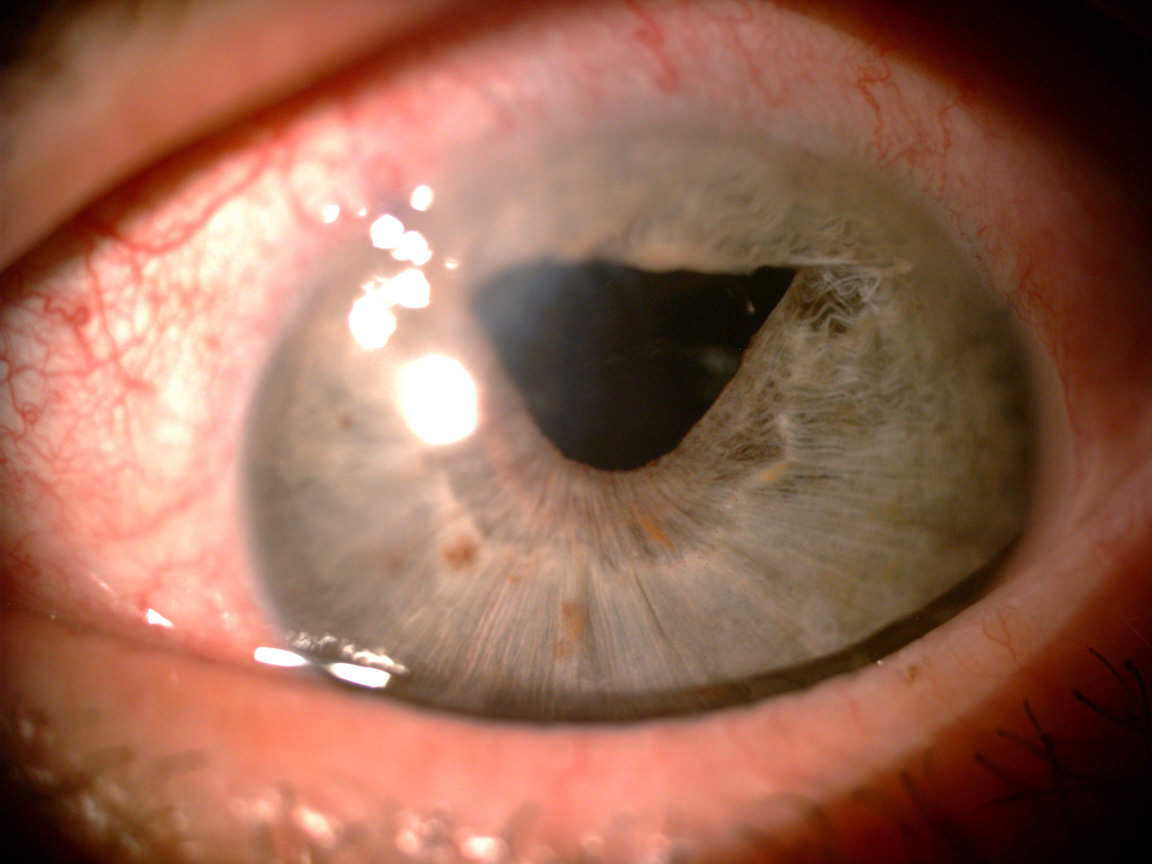}
\caption{Illustration of sample conditions preventing a good quality iris image: blood in the anterior chamber (left), corneal ulcer and distorted pupil (right).}
\label{fig:fkps}
\end{figure}

\paragraph{\textbf{The iris}} \textbf{Rubeosis iridis} is a medical condition of the iris in which new, abnormal blood vessels are found on the surface of the iris. It is usually associated with disease processes in the retina. \textbf{Iris sphincter tear} is a frequent concomitant of both laceration and blunt trauma of the anterior segment. \textbf{Iridodialysis} is defined as a rupture of the iris at its thinnest area, the iris root, manifested as a separation or tearing of the iris from its attachment to the ciliary body. It is usually caused by blunt trauma to the eye but may also be caused by penetrating eye injuries or as a complication of an intraocular surgery. Iridodialyses can often be repaired using suturing techniques. \textbf{Synechiae} are adhesions between the iris and other structures in the eye. \textbf{Iris bombe} occurs when there is a complete adhesion (posterior synechiae) between the iris and the anterior capsule of the lens creating a 360-degree area of adhesion. All of the aforementioned are capable of introducing severe distortions or damage to the iris region.

\paragraph{\textbf{The lens}} \textbf{Anterior lens luxation} (wherein the lens enters the anterior chamber of the eye) can cause damage to the cornea, swelling, and progressive lens opacity, blurring the iris image. \textbf{Phacolytic glaucoma} is an inflammatory glaucoma caused by the leakage of the lens protein through the capsule of a hyper-mature cataract. Escalating \textbf{corneal edema and milky aqueous humor} in the anterior chamber also blur the iris image.

\paragraph{\textbf{Pars plana vitrectomy}} This is a general term used to describe a group of surgical procedures performed in the deeper part of the eye and behind the lens. Silicone oil is used as an intraocular tamponade in the repair of \textbf{retinal detachment} or \textbf{diabetic retinopathy}. Sometimes it may relocate itself to the front of the iris, causing an obstruction that prevents quality iris imaging.

\subsection{Data collection protocol}

For the purpose of this study, a new database was designed and collected specifically for the assessment of how iris recognition is immune or prone to ocular pathologies. The dataset comprises images collected from patients during routine ophthalmological examinations. All patients participating in the study were provided with detailed information on the research and an informed consent has been obtained from each volunteer.

Data collection lasted approximately 16 months. During each visit, both NIR-illuminated images (compliant with the ISO/IEC 19794-6:2011 recommendations) and standard color photographs (for selected cases) were acquired enabling us to perform visual inspection of the illness in samples showing significant alterations to the eye. The data was acquired with three commercial cameras: 1) the IrisGuard AD100 for NIR images, 2) Canon EOS1000D with EF-S 18-55 mm f/3.5-5.6 lens equipped with a Raynox DCR-250 macro converter and a ring flashlight suited for macrophotography, and 3) an ophthalmology slit-lamp camera Topcon DC3, Tab. \ref{table:database_summary}.

\begin{figure*}[!t]
\centering
\includegraphics[width=\textwidth]{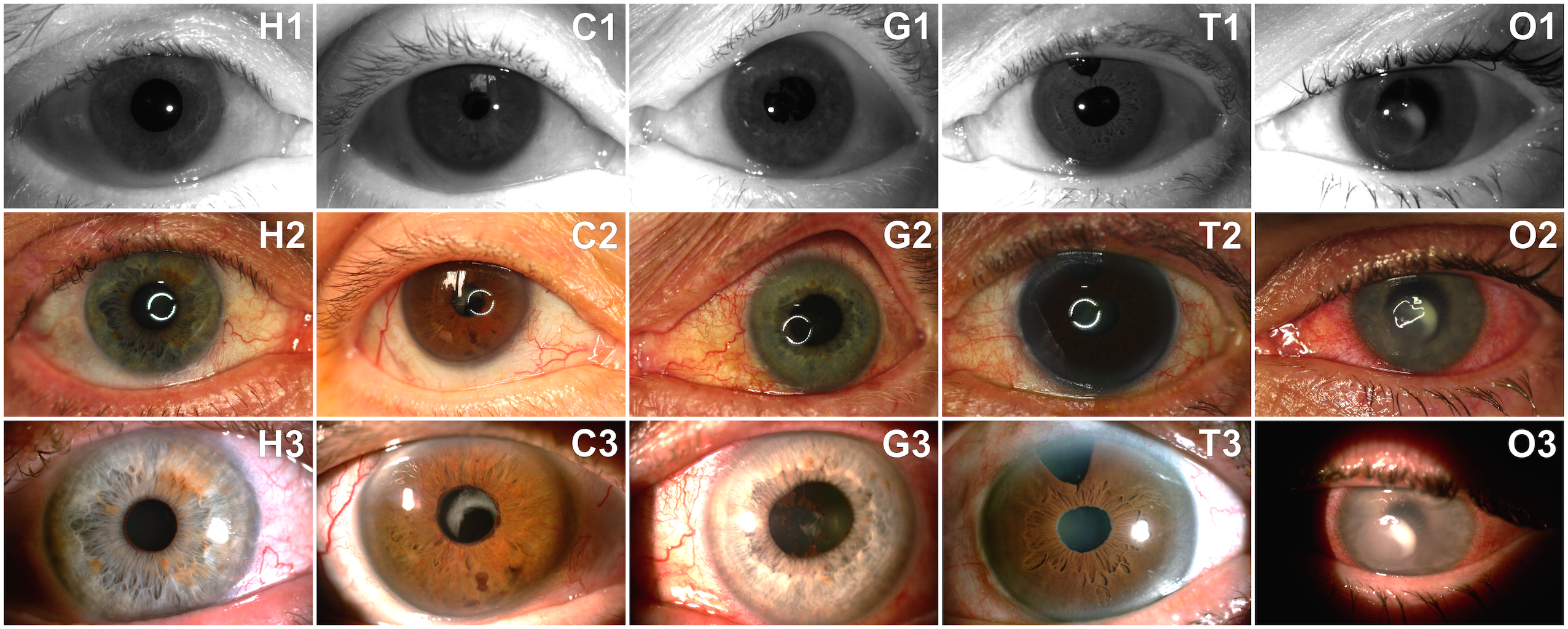}
\caption{Samples of 5 different eyes acquired using three different imaging systems: IrisGuard AD-100 (top row), Canon EOS 1000D (middle row), and Topcon DC3 slit-lamp camera (bottom row). Each column includes samples corresponding to a different group used further in our experimental study, namely: healthy eye (H1-H3), unhealthy eye but with a clearly visible iris pattern (C1-C3), eye with geometrical deviations (G1-G3), eye with iris tissue impairments (T1-T3), and eye with obstructions in front of the iris (O1-O3).}
\label{fig:samples}
\end{figure*}

\subsection{Database statistics} 

The entire dataset comprises 2996 images of 230 distinct irises, Tab. \ref{table:database_summary}. Every class contains NIR-illuminated images, while for some of them visible light photographs were also taken (in cases when visual inspection revealed significant changes to the structures of the eye). Fig. \ref{fig:samples} shows sample images of five different eyes obtained using all three devices. 

Images for 184 irises were captured one acquisition session; for 38 irises, in two sessions; for 6 irises, in three sessions. Finally, for 2 irises, there were four different acquisition sessions. Typically, the second and subsequent sessions contain images obtained after some kind of medical procedure, \eg a cataract surgery. Detailed information, including a precise description of medical conditions and procedures performed in each case, is disclosed in the metadata that accompanies the published dataset. No data censoring was performed when collecting the data, except for immediate removal of images that did not show an eye at all.

\begin{table}[!htb]
\renewcommand{\arraystretch}{1.1}
\caption{Format characterization and numbers of collected images for each sensor.}
\label{table:database_summary}
\centering\footnotesize
\begin{tabular}[t]{|c|c|c|}
\hline
\textbf{Device} & \textbf{Image format} & \textbf{Number of images}\\
\hline
\hline
IrisGuard AD100 & grayscale, 640x480 BMP & 1793\\
\hline
Canon EOS1000D & color, 10 Mpixel JPEG & 868\\
\hline
Topcon DC3 & color, 8 Mpixel
 JPEG & 335\\
\hline
\end{tabular}
\end{table}

\subsection{Disease representation in the database and general population prevalence} 

To gain some insight on the prevalence of eye disease in the general population, some preliminary classification must be made. While the database images represent more than 20 different medical conditions, most of them can be roughly classified into three main groups of related disorders or conditions and their effects on the eye:

\begin{enumerate}
	\item Cataract and related conditions, which can be found in different stages of the treatment process:
	\begin{itemize}
		\item opacified lens,
		\item lens implant after the surgery,
		\item aphakia -- lack of lens after complicated surgeries,
		\item capsulotomies -- incisions in the lens to correct capsule opacity.
	\end{itemize}
	\item Glaucoma and associated medical procedures:
	\begin{itemize}
		\item trabeculectomy -- surgical removal of a part of the upper portion of the iris tissue, 
		\item iridotomy -- puncturing the iris using laser radiation.
	\end{itemize} 
	\item Cornea pathologies leading to opacification, making it more difficult to obtain a good image of the iris beneath it:
	\begin{itemize}
		\item as a result of corneal inflammation and ulcers,
		\item as a result of trauma and chemical burns.
	\end{itemize}
\end{enumerate} 

In 2012, 24 million Americans suffered from cataract (estimated to rise to 38.7 million in 2030) and 2.7 million suffered from glaucoma (4.3 million in 2030) \cite{VisionProblemsReport}. In 2014, over 79 thousand corneal transplant surgeries were performed \cite{CorneaReport}. Such high volumes of pathology occurrences (especially for cataract, expected to affect over 10\% of the US population by 2030) make it crucial to assess iris recognition performance under in the presence of common pathologies.

The exact description of each medical case represented in the database can be found in the medical description file included with this publication as Supplementary Material. All interested researchers are encouraged to download the additional data in order to carry out their own interpretations and analyses with this data.

\subsection{Access to the database} 

The database collected by the authors as a part of this study is available to all interested researchers for non-commercial research applications. Further information on how to get the data can be found at: \url{http://zbum.ia.pw.edu.pl/EN}$\rightarrow$Research$\rightarrow$Databases

\section{Experimental study}
\label{section:experimental}

\subsection{Preparing and dividing the data} 

Our data shows that most unhealthy eyes suffer from more than one condition, often unrelated and impacting the eye in different ways. While some illnesses cause the pupil to distort and deviate from its usual circular shape, other pathologies impact the iris directly or cause changes to other parts of the eyeball, such as the uvea, the cornea, the anterior chamber, or even the retina. Hence, conducting an insightful analysis and, in particular, a separate analysis for each individual impairment, may be challenging or even impossible. The data was, therefore, partitioned respectively according to the \emph{type} of influence that a given ocular pathology inflicts on the eye. This allowed us to devise five different partitions: \emph{Healthy}, comprising healthy eyes only; \emph{Clear}, made up of eyes with a disease present, but having no perceivable effect on the eye structures; \emph{Geometry} (eyes whose pupil geometry has been distorted by the pathology); \emph{Tissue} (eyes with damage inflicted on the iris tissue) and \emph{Obstructions}, encompassing the eyes with obstructions present in front of the iris. Figure \ref{fig:samples} shows sample images belonging to each partition. Table \ref{table:subsets} presents the numbers of classes (\ie different eyes) and images in each partition.

For the purpose of this study, we have selected a subset of the original dataset that comprises only NIR images obtained during the first acquisition session for each eye. Images representing eyes to which pupil dilating drugs were administered have also been excluded from the dataset. This was done in order to leverage disease-induced changes only, and to eliminate any changes caused by the examiner’s actions during the patient’s visit or the effects of treatment that patients may have undergone between individual visits. For this reason, the number of distinct irises and images (shown in Tab. \ref{table:subsets}) do not total 230 (the number of all unique irises in the dataset) and 1793 (the number of all NIR iris images in the dataset), respectively.

\begin{table*}[!ht]
\renewcommand{\arraystretch}{1.1}
\caption{Numbers of unique irises (classes) and numbers of unique samples in each of the five data partitions, and the total number of classes and samples in the data subset selected for this particular study.}
\label{table:subsets}
\centering\footnotesize
\begin{tabular}[t]{|c|c|c|}
\hline
\textbf{Data partition} & \textbf{Number of irises} & \textbf{Number of images} \\
\hline
\hline
\textbf{Healthy} & 35 & 216 \\
\hline
\textbf{Clear} & 87 & 568 \\
\hline
\textbf{Geometry} & 53 & 312 \\
\hline
\textbf{Tissue} & 8 & 50 \\
\hline
\textbf{Obstructions} & 36 & 207 \\
\hline\hline
\textbf{Total} & \textbf{219} & \textbf{1353} \\
\hline
\end{tabular}
\end{table*}

\subsection{Evaluation methodology} 
\label{sec:methodology}

\textbf{To answer the first question} formulated in the introduction, failure-to-enroll error rates (FTE) are calculated for each database partition and using four different iris recognition methods.

\textbf{To answer questions two and three}, all possible genuine comparison scores were generated and full cross-comparisons were executed to obtain all possible impostor comparison scores for each dataset partition. To judge whether the observed differences in comparison scores across partitions can be considered as samples drawn from the same distribution, a two-sample Kolmogorov-Smirnov test is applied with the significance level $\alpha=0.05$ (further referred to as K-S test). The K-S test makes no assumptions on the distributions (apart from their continuity) and the test statistics simply quantifies the distance between two empirical cumulative distribution functions $F(x_1)$ and $F(x_2)$ of the random variables $x_1, x_2$ being compared.

The generation of all possible genuine and impostor comparisons delivers the richest information about the population under consideration. At the same time, it inevitably introduces statistical dependencies among samples, narrowing the degrees of freedom in statistical testing. To mitigate this problem, we resample (with replacement) each set of comparison scores 1,000 times for genuine scores and 10,000 times for impostor scores to end up with statistically independent samples used further in K-S testing. These analyses were done independently for four iris recognition methods, following the procedure used to answer the first question.

Finally, \textbf{to answer the fourth question} regarding segmentation outcomes, an analysis of segmentation errors and visual inspection of eye samples resulting in the worst comparison scores were performed for selected matchers.

\subsection{Iris recognition methods} 
\label{methods}

In this work four different, independent iris recognition methods were employed. The first comes from an academic community, while the three remaining algorithms are widely recognized and well-regarded commercially available products. All four methods are briefly characterized in this subsection.

\textbf{OSIRIS} ({\it Open Source for IRIS}) \cite{OSIRIS} is an open source implementation of the well-known Daugman’s iris recognition concept. The OSIRIS software comprises four independent operations: a) image segmentation employing the Viterbi algorithm, b) image normalization based on a rubber-sheet model, c) iris coding by quantization of the Gabor filtering outcomes, and d) iris code comparison based on fractional Hamming distance. Original implementation (used in this work) calculates the iris code for three different resolutions of the complex Gabor filter kernel. Phase of only 256 unique and equidistantly located points is quantized to one of four possible quadrants of the complex plane (employing only two bits per point). This results in the iris code length of 1536 bits (3 resolutions $\times$ 256 points $\times$ 2 bits for the coding point’s phase). The occlusion mask (calculated at the segmentation stage) eliminates those iris code bits that correspond to non-iris areas. As in Daugman’s solution, we should expect a dissimilarity score close to zero when comparing samples of the same eye, and close to 0.5 when comparing different irises (as in the comparison of two sequences of heads and tails obtained in independent coin tosses). Due to rotation compensation typically realized by shifting the iris code and finding the best match, the distribution of impostor comparison scores is typically skewed towards smaller values (about 0.4).

\textbf{MIRLIN} ({\it Monro Iris Recognition Library}) has been offered on the market as a Software Development Kit (SDK) \cite{MIRLIN}. It employs a discrete cosine transform calculated for local iris image patches to deliver the binary iris code \cite{Monro2007}. Similarly to Daugman's approach, the iris codes are compared to each other by calculating the fractional Hamming distance, normalized by the number of valid iris code bits, \ie originating from non-occluded iris regions. As for the OSIRIS method, comparing two images of the same eye should yield a fractional Hamming distance close to zero, while the distance for two different eyes should oscillate around 0.5. The advantage of MIRLIN is the visualization of automatic segmentation results, helpful when analyzing sources of possible errors when processing images of unhealthy eyes.

\textbf{VeriEye}, the third matcher involved in this study, is another commercial product offered by Neurotechnology for more than a decade \cite{VeriEye}. It incorporates an unpublished iris encoding methodology, although thoroughly evaluated in numerous applications and scientific projects, \eg in NIST ICE 2005 project \cite{ICE2005}. The manufacturer claims to employ an off-axis iris localization with the use of active shape modeling. In contrast to OSIRIS and MIRLIN, the VeriEye delivers a similarity score (but not the Hamming distance) between two iris images. The higher the score, the more similar the images. A zero score denotes a perfect non-match.

\textbf{IriCore} is the fourth iris recognition system employed in this study and offered on the market as an SDK by IriTech Inc. \cite{IriCore}. As with the VeriEye solution, scientific papers divulging the implemented methodology have not been published. Nonetheless, the IriCore implementation was placed in a narrow set of the best solutions tested by NIST in 2005 \cite{ICE2005}. The manufacturer claims conformance with two editions of the ISO/IEC 19794-6 standard (one issued in 2005 and the most current one published in 2011). Using the IriCore system to compare two same-eye images should result in a near-zero dissimilarity score, while scores between 1.1 and 2.0 are typically observed when comparing images of two different eyes.

\section{Results}
\label{experiments}
\subsection{Enrollment performance (re: Question 1)} 

FTE rates obtained in each partition (Tab. \ref{table:FTE}) suggest that iris recognition performs particularly poorly for samples included in the \emph{Geometry} and \emph{Obstructions} partitions. Those partitions comprise images in which the pupil is either distorted or not visible at all due to various types of occlusions. It is worth noting that the enrollment process realized for different iris recognition methods is affected unevenly across the methods. This can be explained by different quality metrics implemented in the employed algorithms and their varying sensitivity to quality issues generated by eye disorders. Hence, \textbf{the answer to Question 1 is that the enrollment stage is sensitive to those conditions that distort pupil geometry or that obstruct, partially or completely, the iris pattern. The observed impact on the enrollment process is uneven across algorithms}.

\begin{table*}[!ht]
\renewcommand{\arraystretch}{1.1}
\caption{FTE rates obtained in each partition for four iris recognition methods used in this work. The worst result for each method is in \textbf{bold type}. The second column provides reference to sample images from each subset, shown in Fig. \ref{fig:samples}}
\label{table:FTE}
\centering\footnotesize
\begin{tabular}[t]{|c|c|c|c|c|c|}
\hline
Partition & Samples in Fig. \ref{fig:samples} & {\bf MIRLIN} & {\bf VeriEye} & {\bf OSIRIS} & {\bf IriCore}\\
\hline
\hline
\textbf{Healthy} & H1 - H3 & 1.85\% & 0\% & 0\% & 0\% \\
\hline
\textbf{Clear} & C1 - C3 & 4.40\%  & 0	\% & 1.23\% & 0\%\\
\hline
\textbf{Geometry} & G1 - G3 & 16.03\% & \textbf{5.13\%} & 5.45\% & 0.32\%\\
\hline
\textbf{Tissue} & T1 - T3 & 2\% & 0\% & 0\% & 0\%\\
\hline
\textbf{Obstructions} & O1 - O3 & \textbf{18.36\%}  &  3.86\% & \textbf{8.21\%} & \textbf{0.97\%}\\
\hline
\end{tabular}
\end{table*}

\subsection{Matching performance (re: Questions 2 and 3)}

Cumulative distributions $F$ of the all possible comparison scores calculated for four iris recognition algorithms are shown in Figs. \ref{fig:cdfML} -- \ref{fig:cdfIC}. $F$ graphs were selected intentionally instead of ROC curves to highlight the differences in genuine and impostor scores independently. In each figure we collectively plot five $F$ graphs (for all considered partitions) along with the mean values of the comparison scores to visualize possible differences among partitions. These bring the first observation that eye disorders impact same-eye comparisons to a higher extent when compared to different-eye matching, since differences between $F$ graphs for genuine comparisons are significantly larger than for impostor comparisons.

\begin{figure}[!htb]
\centering
\includegraphics[width=0.48\textwidth]{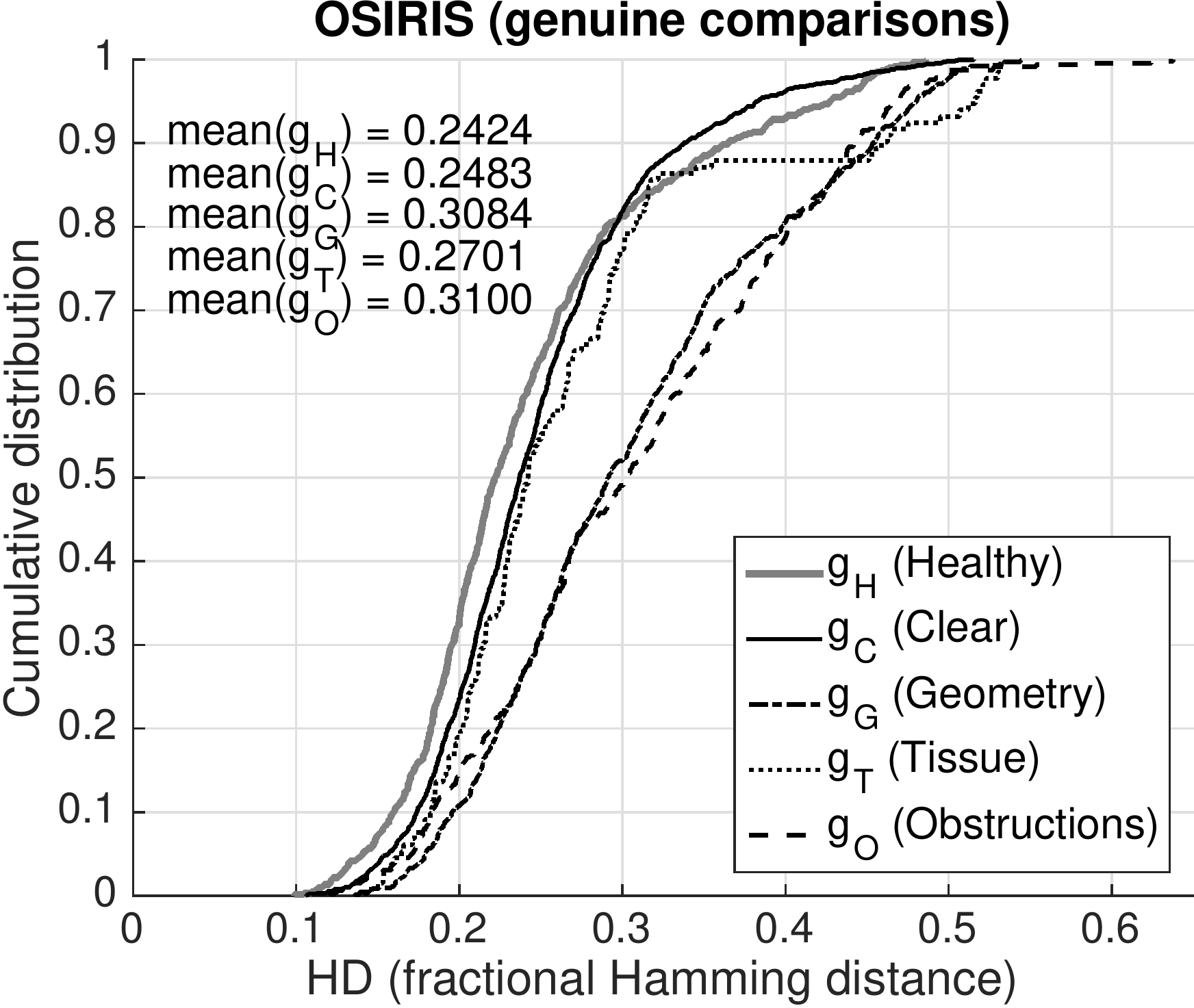}\hfill
\includegraphics[width=0.48\textwidth]{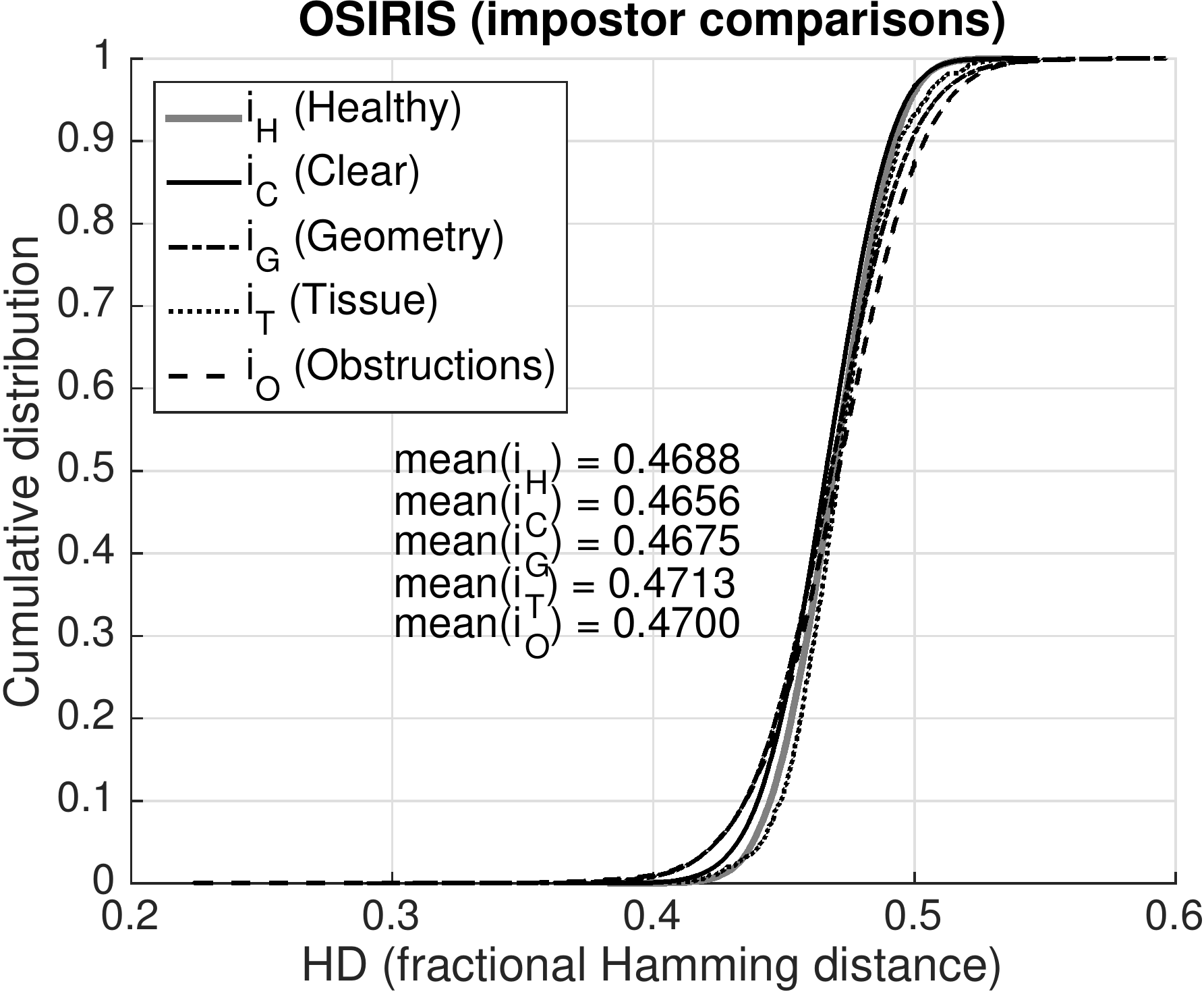}
\caption{Cumulative distribution functions $F$ of genuine scores {\bf (left)} and impostor scores {\bf (right)} for the {\bf OSIRIS} method. Mean values of comparison scores for each partition are also presented.}
\label{fig:cdfOS}
\end{figure}

\begin{figure}[!htb]
\centering
\includegraphics[width=0.48\textwidth]{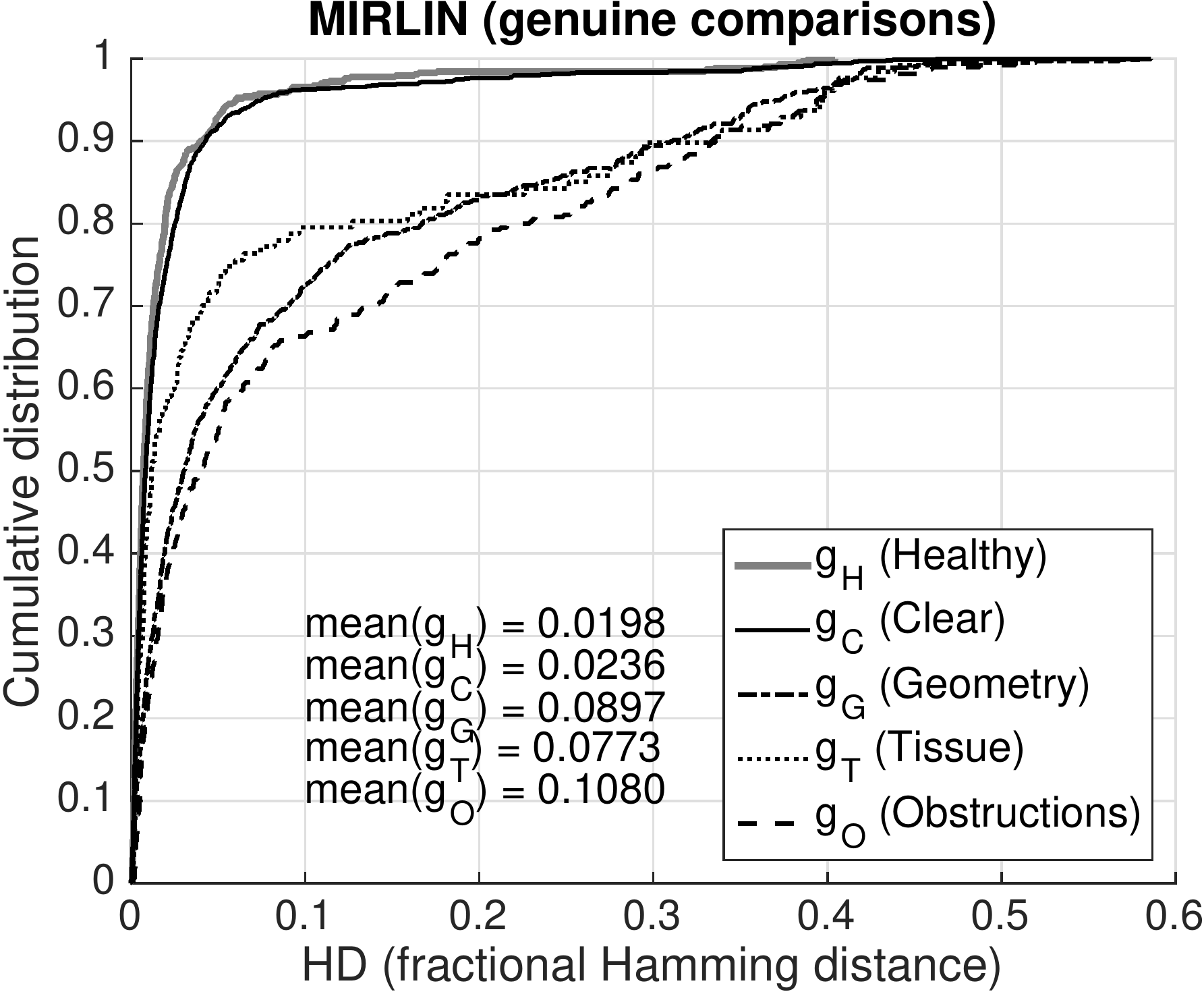}\hfill
\includegraphics[width=0.48\textwidth]{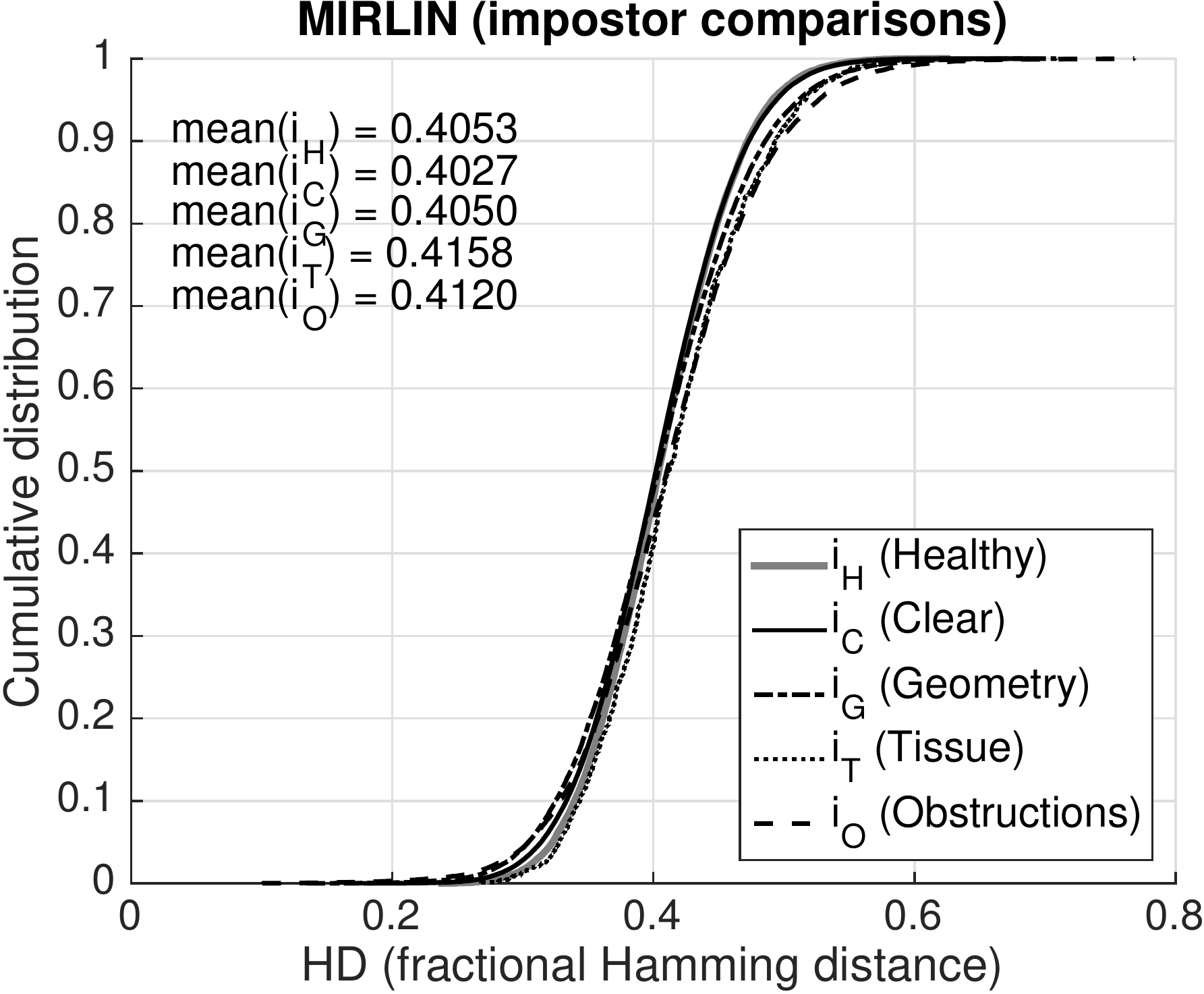}
\caption{Same as in Fig. \ref{fig:cdfOS}, except that the {\bf MIRLIN} method was employed to generate scores.}
\label{fig:cdfML}
\end{figure}

\begin{figure}[!htb]
\centering
\includegraphics[width=0.48\textwidth]{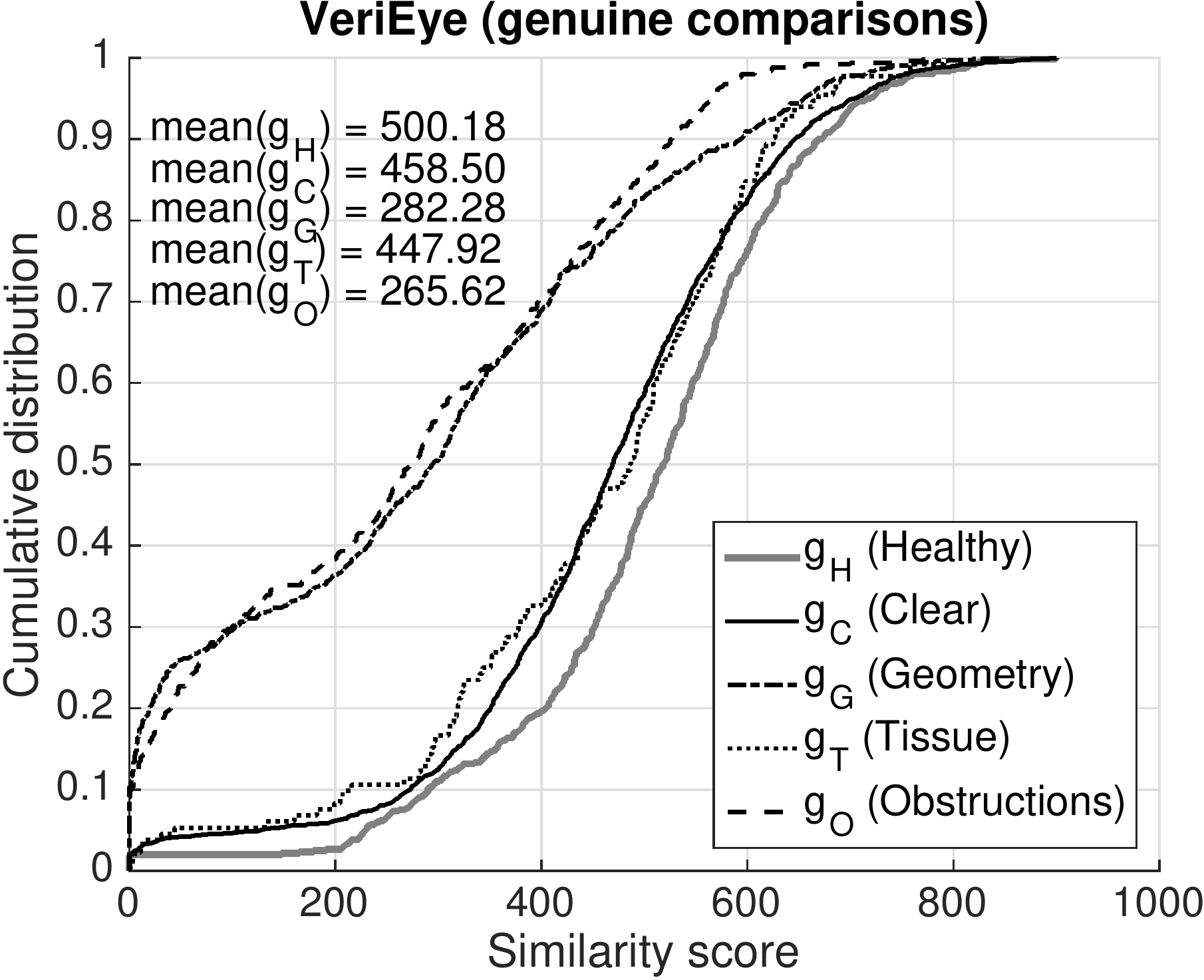}\hfill
\includegraphics[width=0.48\textwidth]{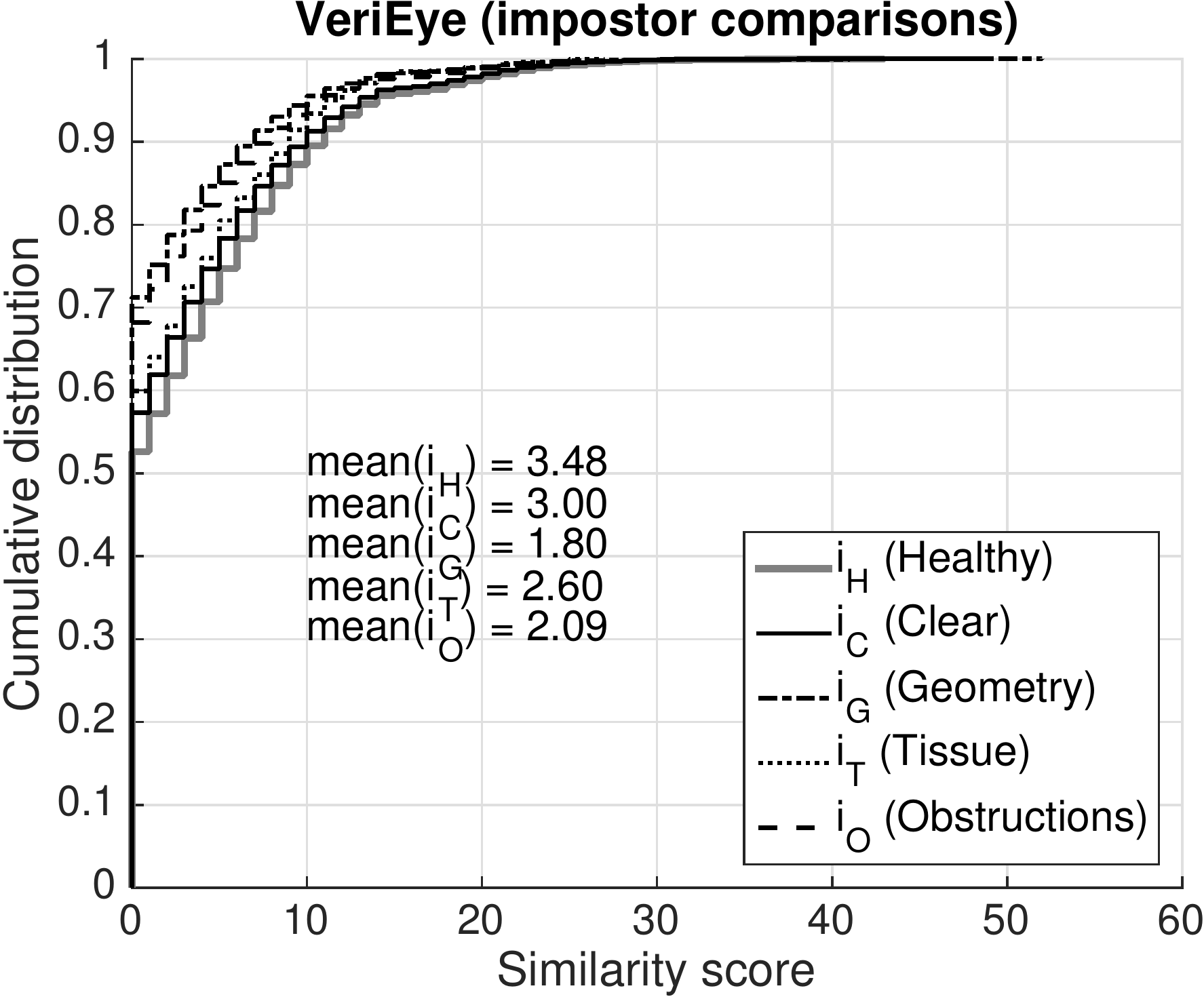}
\caption{Same as in Fig. \ref{fig:cdfOS}, except that the {\bf VeriEye} method was employed to generate scores.}
\label{fig:cdfNT}
\end{figure}

\begin{figure}[!htb]
\centering
\includegraphics[width=0.48\textwidth]{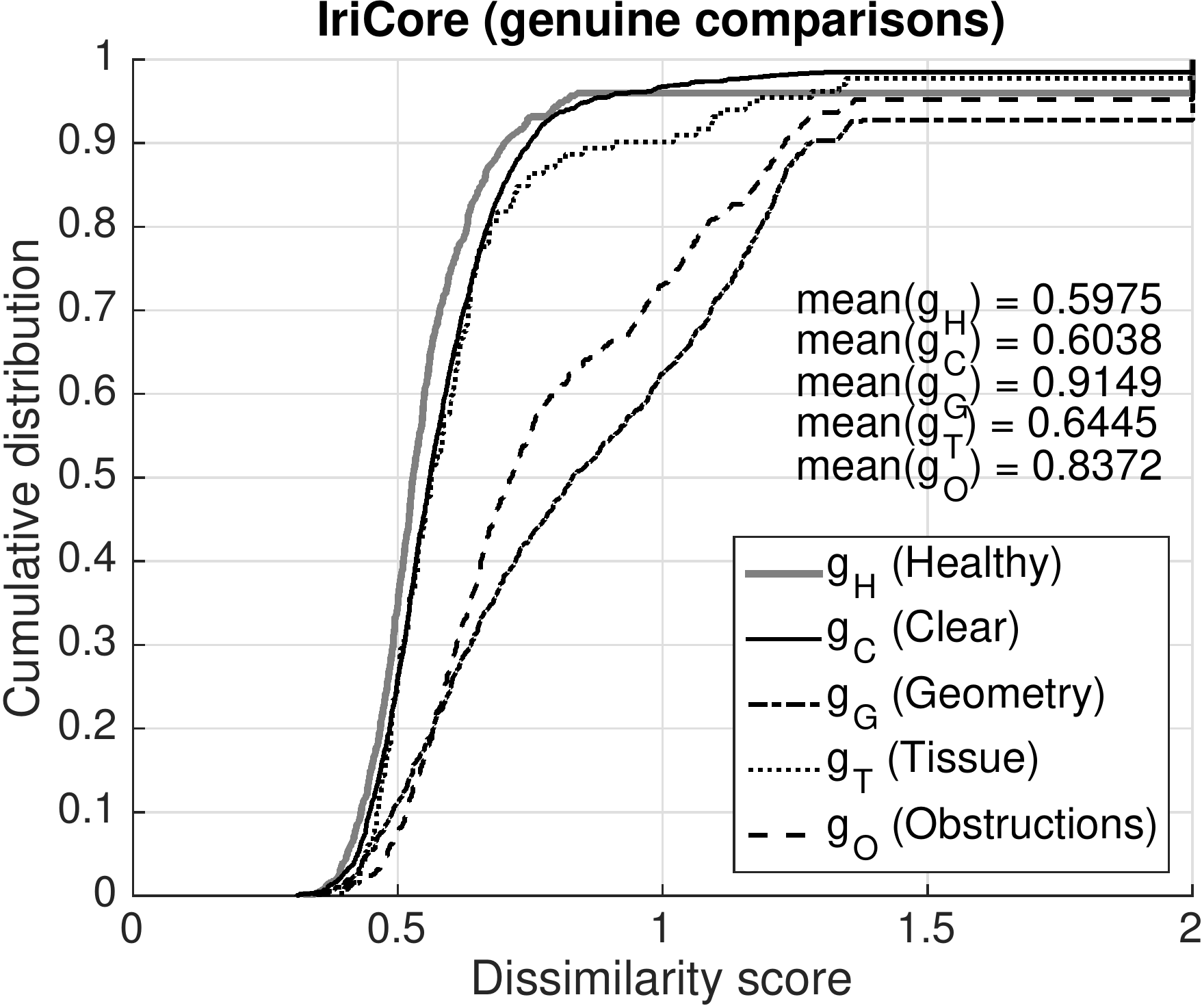}\hfill
\includegraphics[width=0.48\textwidth]{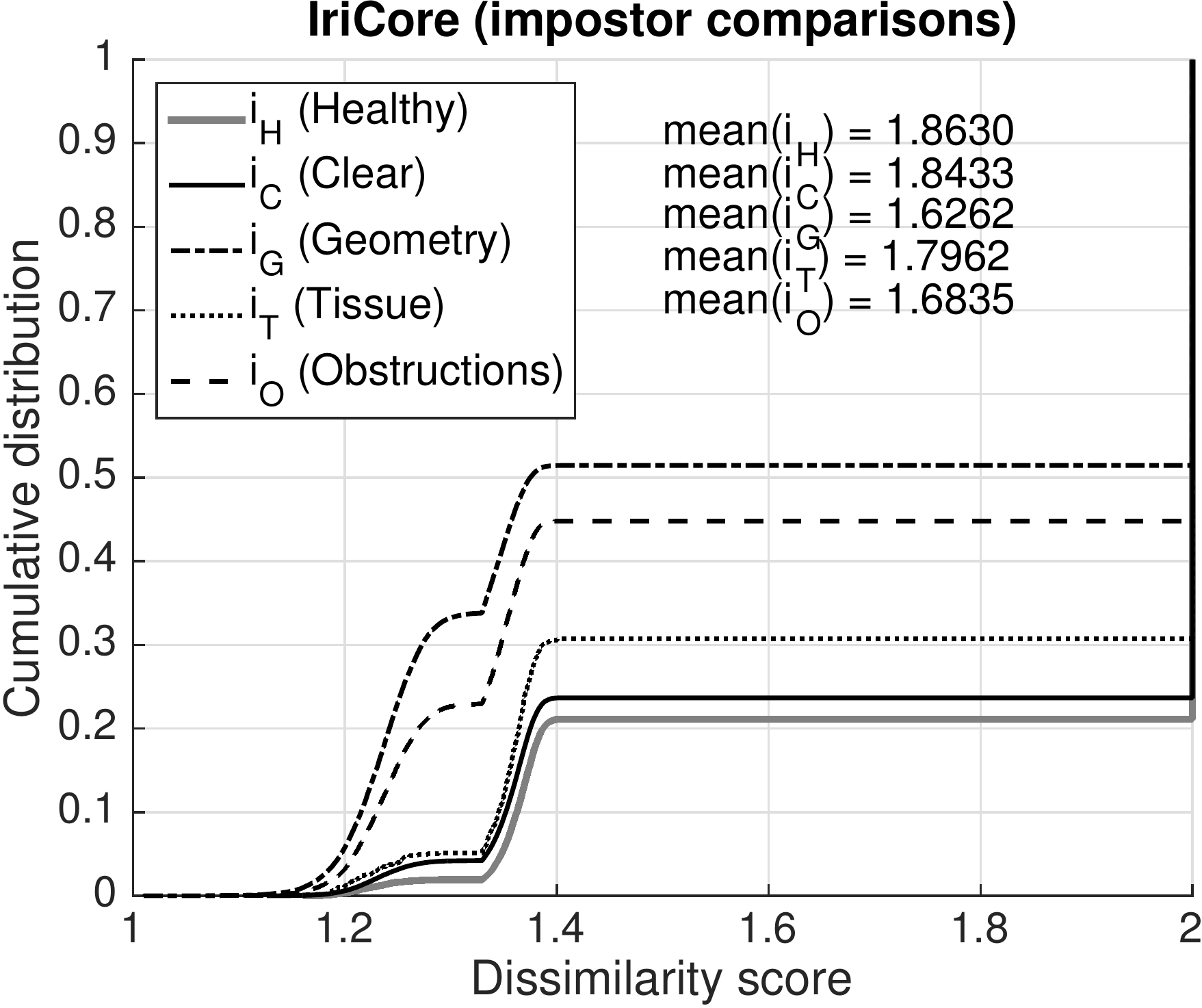}
\caption{Same as in Fig. \ref{fig:cdfOS}, except that the {\bf IriCore} method was employed to generate scores.}
\label{fig:cdfIC}
\end{figure}

Differences observed when visually inspecting the $F$ graphs and analyzing the corresponding mean values should, however, be confirmed by statistical testing. As noted in Sec. \ref{sec:methodology}, the Kolmogorov-Smirnov testing at the significance level $\alpha=0.05$ is employed for this purpose.

The null hypotheses $H_0$ in all tests state that samples originating from two compared partitions $x_1, x_2$ are drawn from the same distribution, \ie $F(x_1)$ is equivalent to $F(x_2)$. However, the alternative hypotheses differ depending on whether $x_1$ and $x_2$ represent genuine or impostor comparison scores. For same-eye comparisons we expect to obtain better scores within \emph{Healthy} samples when compared to samples of the remaining partitions. \emph{Better} scores mean \emph{lower} scores for MIRLIN, OSIRIS and IriCore methods. Hence, in the case of these matchers the alternative hypothesis $H_1$ is $F(x_1) > F(x_2)$, where $x_1$ corresponds to the \emph{Healthy} genuine scores, and $x_2$ corresponds to the genuine scores calculated for the remaining partitions. It means that for lower (\ie better) scores the corresponding $F$ graph is above the graph $F$ calculated for higher (\ie worse) scores. For VeriEye matcher, \emph{better} scores mean \emph{higher} scores. Thus, in this case the alternative hypothesis $H_1$ is $F(x_1) < F(x_2)$, with the same meaning of $x_1$ and $x_2$ as above. Consequently, a one-sided two-sample K-S test is applied in case of genuine scores. For different-eye comparisons we do not make any specific suppositions on the relation between comparisons scores obtained in different partitions. Thus, the alternative hypothesis $H_1$ is $F(x_1) \nsim F(x_2)$, where $x_1$ corresponds to the \emph{Healthy} impostor scores, and $x_2$ corresponds to the impostor scores calculated for the remaining partitions. It means that a two-sided two-sample K-S test is applied when analyzing impostor scores. Tables \ref{table:statTestsGenuine} and \ref{table:statTestsImpostor} present summary of statistical testing for genuine and impostor scores (resampled with replacement, as described in Sec. \ref{sec:methodology}), respectively; $p$-values less than $\alpha=0.05$ denote that the null hypothesis $H_0$ was rejected and the corresponding alternative hypothesis $H_1$ was selected.

\begin{table}[!htb]
\renewcommand{\arraystretch}{1.1}
\caption{Summary of statistical testing of differences in distributions of genuine comparison scores obtained for all four iris recognition methods in five partitions of the dataset using \textbf{Kolmogorov-Smirnov test} applied for resampled data (with replacement). The null hypotheses $H_0$ in all tests state that the samples originating from two compared partitions are drawn from the same distribution. Alternative hypotheses are detailed in rows labeled $H_1$ (note using one-sided test). $F(g_k)$ denotes the cumulative distribution function of $g_k$, where $g_k$ denotes the genuine scores calculated to $k$-th partition.}
\label{table:statTestsGenuine}
\centering\scriptsize
\begin{tabular}[t]{cc|c|c|c|c|}
\cline{3-6}
& & \emph{Clear} ($g_c$) & \emph{Geometry} ($g_g$) & \emph{Tissue} ($g_t$) & \emph{Obstructions} ($g_o$) \\
& & vs. \emph{Healthy} ($g_h$) & vs. \emph{Healthy} ($g_h$) & vs. \emph{Healthy} ($g_h$) & vs. \emph{Healthy} ($g_h$) \\\hline\hline
\multicolumn{1}{|c}{\multirow{2}{*}{\bf OSIRIS}} & \multicolumn{1}{|c|}{$H_1$} & $F(g_c) < F(g_h)$ & $F(g_g) < F(g_h)$ & $F(g_t) < F(g_h)$ & $F(g_o) < F(g_h)$  \\\cline{2-6}
\multicolumn{1}{|c}{} & \multicolumn{1}{|c|}{$p$-value} & $< 0.0001$  & \textapprox 0 & $< 0.0001$  & \textapprox 0 \\\hline
\multicolumn{1}{|c}{\multirow{2}{*}{\bf MIRLIN}} & \multicolumn{1}{|c|}{$H_1$} & $F(g_c) < F(g_h)$ & $F(g_g) < F(g_h)$ & $F(g_t)<F(g_h)$ & $F(g_o)<F(g_h)$  \\\cline{2-6}
\multicolumn{1}{|c}{} & \multicolumn{1}{|c|}{$p$-value} & $< 0.0001$  & \textapprox 0 & \textapprox 0 & \textapprox 0 \\\hline
\multicolumn{1}{|c}{\multirow{2}{*}{\bf VeriEye}} & \multicolumn{1}{|c|}{$H_1$} & $F(g_c) > F(g_h)$ & $F(g_g) > F(g_h)$ & $F(g_t) > F(g_h)$ & $F(g_o) > F(g_h)$  \\\cline{2-6}
\multicolumn{1}{|c}{} & \multicolumn{1}{|c|}{$p$-value} & $< 0.0001$ & \textapprox 0 & $< 0.0001$ & \textapprox 0 \\\hline
\multicolumn{1}{|c}{\multirow{2}{*}{\bf IriCore}} & \multicolumn{1}{|c|}{$H_1$} & $F(g_c) < F(g_h)$ & $F(g_g) < F(g_h)$ & $F(g_t)<(g_h)$ & $F(g_o)<F(g_h)$  \\\cline{2-6}
\multicolumn{1}{|c}{} & \multicolumn{1}{|c|}{$p$-value} & $< 0.0001$ & \textapprox 0 & $< 0.0001$ & \textapprox 0 \\\hline
\end{tabular}
\end{table}

\begin{table}[!htb]
\renewcommand{\arraystretch}{1.1}
\caption{Same as in Table \ref{table:statTestsGenuine}, except that impostor comparison scores are analyzed and two-sided Kolmogorov-Smirnov test (for the resampled data) was applied. $F(i_k)$ denotes the cumulative distribution function of $i_k$, where $i_k$ denotes the impostor scores calculated to $k$-th partition.}
\label{table:statTestsImpostor}
\centering\scriptsize
\begin{tabular}[t]{c|c|c|c|c|}
\cline{2-5}
& \emph{Clear} ($i_c$) & \emph{Geometry} ($i_g$) & \emph{Tissue} ($i_t$) & \emph{Obstructions} ($i_o$) \\
& vs. \emph{Healthy} ($i_h$) & vs. \emph{Healthy} ($i_h$) & vs. \emph{Healthy} ($i_h$) & vs. \emph{Healthy} ($i_h$) \\
& $H_1: F(i_c) \nsim F(i_h)$ & $H_1: F(i_g) \nsim F(i_h)$ & $H_1: F(i_t) \nsim F(i_h)$ &  $H_1: F(i_o) \nsim F(i_h)$\\\hline\hline
\multicolumn{1}{|c|}{\bf OSIRIS} & $< 0.0001$ & \textapprox 0 & $< 0.0001$ & \textapprox 0 \\\hline
\multicolumn{1}{|c|}{\bf MIRLIN} &  $< 0.0001$ & $< 0.0001$ & \textapprox 0 & \textapprox 0 \\\hline
\multicolumn{1}{|c|}{\bf VeriEye} & $< 0.0001$ & \textapprox 0 & \textapprox 0 & \textapprox 0 \\\hline
\multicolumn{1}{|c|}{\bf IriCore} & $< 0.0001$ & \textapprox 0 & \textapprox 0  & \textapprox 0 \\\hline
\end{tabular}
\end{table}

Having both the $F$ graphs and results of the statistical testing, let us refer to our original questions. To verify whether iris recognition performs worse for unhealthy eyes (even those that do not reveal visible changes) than healthy eyes (Question 2), genuine and impostor scores for \emph{Healthy} and \emph{Clear} eyes are compared for those irises that were correctly enrolled (solid lines in Figs. \ref{fig:cdfOS} -- \ref{fig:cdfIC}). For all iris coding methodologies, both the cumulative distribution functions and mean genuine comparison scores suggest that the performance in \emph{Clear} partition is slightly worse than for \emph{Healthy} eyes. Impostor scores differ to a lesser extent. These small differences are, however, statistically significant, since $p$-value $<\alpha=0.05$ for all K-S tests (cf. first columns of Tables \ref{table:statTestsGenuine} and \ref{table:statTestsImpostor} for genuine and impostor comparisons, respectively). Therefore, \textbf{the answer to Question 2 is affirmative, but the observed differences are uneven across the methods and small for impostor comparison scores}. Additionally, the iris recognition method based on Discrete Cosine Transform (MIRLIN) seems to be more robust to those eye conditions that cause no visible changes in the iris structure.

To answer the third question, related to how diseases introducing visible structural changes to the eye impact the performance, the genuine and impostor comparison scores were calculated in the following three partitions of unhealthy eyes: \emph{Geometry}, \emph{Tissue} and \emph{Obstructions} (dashed and dotted lines in Figs. \ref{fig:cdfOS} -- \ref{fig:cdfIC}). In this experiment, a serious deterioration in within-class matching performance can be observed, and these differences are statistically significant according to the K-S test (cf. three last columns of Tables 4 and 5 for genuine and impostor scores, respectively). Conditions that generate obstructions of the iris structure have the most influence on iris recognition methods.  Observing the mean values of the genuine comparison scores for all of the applied iris recognition methods, it is evident that such a large shift in genuine comparison distributions would have a significant influence on a false-non match rate, difficult to  compensate for  by modification of the acceptance threshold. However, observed shifts of the impostor score distributions are small and have rather marginal influence on the overall system performance. \textbf{Therefore, in answer to Question 3, we state that all eye conditions resulting in visible eye structure impairments have a substantial influence on within-class variability. The largest increase in the probability of a false non-match is expected for those conditions that create geometrical deformations of the pupillary area and introduce obstructions of the iris}.

\subsection{Sources of errors (re: Question 4)} 

The most likely cause of errors is a failed segmentation resulting in encoding portions of images unrelated to the iris. To assess whether this is true, in this study we carefully inspected sample pairs yielding genuine match scores below typical acceptance thresholds (\ie $0.32$ for MIRLIN and OSIRIS, the two algorithms that were used to look into the segmentation results). This visual examination confirmed that segmentation errors are the most prevalent source of decrease in recognition accuracy. These errors were, for the most part, caused by the pupil segmentation algorithms which misinterpreted an irregular pupil boundary (corneal occlusions that obstruct the pupil and the iris) or damage to the iris tissue as the pupil itself. In particular, all of the sample pairs generating the worst OSIRIS scores (\emph{Geometry} and \emph{Obstructions}) derive their poor performance from the segmentation errors. A similar result occurs when the MIRLIN matcher is involved, except for two images that produce poor scores in the \emph{Clear} subset: one blurred, and the other showing no discernible flaw.

We were unable to find a way to read segmentation outcomes for the VeriEye and IriCore algorithms. However, when examining those samples that perform the worst when using this method, one may easily identify conditions responsible for errors, namely: significant geometrical distortions, severe corneal hazes, blurred boundary between the iris and the pupil. Thus, segmentation failures may be the case here as well.

\section{Conclusions} 
\label{conclusions}

This paper presents the most thorough and comprehensive analysis aimed at explaining one of the most elusive and unexplored aspects of iris recognition; that is, how iris recognition methods perform in the presence of ophthalmic disorders. Our research required a new approach to data acquisition and analysis, as most patients tend to suffer not from one, but often from several, sometimes unrelated, ocular pathologies. That being said, we applied data categorization related to the type of impact or damage afflicting the eye, instead of relating it to a medical disease taxonomy. Following that, four independent iris recognition algorithms were used to help understand this phenomenon.

Deterioration in recognition performance begins manifesting itself as early as at the enrollment stage, with FTE rates being significantly higher for eyes with geometrical distortions in the pupillary area, or those with pathology induced objects interfering with correct imaging of the iris (obstructions). Those two types of impairments also have a profound effect on the reliability of the comparison stage, since such eyes perform significantly worse (mostly for same-eye comparisons) than do their healthy counterparts (as confirmed by the Kolmogorov-Smirnov test). For all algorithms employed, there are also slight changes in comparison scores obtained from healthy eyes and in those afflicted with, but not visibly affected by, disease. While the observed differences are statistically significant, they are rather small and do not show a potential for generating false rejections or false acceptances.

Erroneous execution of automatic iris image segmentation seems to be the most plausible candidate for causing this drop in performance. An incorrectly localized iris characterizes the worst performing images, leading to encoding of data that don’t necessarily represent iris tissue.

To guard against the impact that medical conditions may have on recognition accuracy, we would advise performing visual inspection of the eye condition (by a person supervising enrollment) since the diseases and pathologies most likely to cause a notable drop in the performance of the algorithms are also the most likely to affect the eye in a visually perceivable manner.

Due to the inherent complexity of the subject, more research is necessary to fully comprehend its nature and consequences. Results presented in this paper, however, should certainly not be seen as the basis for discrediting iris recognition as a method for biometric authentication, nor should they cast doubt that any of the employed algorithms can be deployed. In case of severe ocular pathology, a degradation in iris recognition accuracy, regardless of the methodology employed, is to be expected. In this study, we offer experimental evidence to better inform the biometric community on these issues so that adequate countermeasures may be employed. In this way, iris recognition may continue to serve as a fast, safe, and secure biometric method.

\bibliographystyle{model1-num-names}

\section*{\refname}
\small
\bibliography{refs}

\end{document}